\documentclass[letterpaper, 10 pt, conference]{ieeeconf}  

\IEEEoverridecommandlockouts                              

\usepackage{graphics} 
\usepackage{epsfig} 
\usepackage{mathptmx} 
\usepackage{times} 
\usepackage{amsmath} 
\usepackage{amssymb}  
\usepackage{color}
\usepackage{url}
\usepackage{bm}

\usepackage{subcaption}
\let\labelindent\relax
\usepackage[inline]{enumitem}

\title{\LARGE \bf
Signs of Language: Embodied Sign Language Fingerspelling Acquisition from Demonstrations for Human-Robot Interaction
}

\author{Federico Tavella$^{1, 2, 3}$ and Aphrodite Galata$^{1}$ and Angelo Cangelosi$^{1, 2}$
\thanks{$^{1}$Centre for Robotics and AI, University of Manchester, UK}
\thanks{$^{2}$Department of Computer Science, University of Manchester, UK}%
\thanks{$^{3}$Corresponding author
        {\tt\footnotesize federico.tavella@manchester.ac.uk}}%
\thanks{This work was supported by the UKRI Trustworthy Autonomous Systems Node in Trust(EP/V026682/1)}
}

\begin{document}

\maketitle
\thispagestyle{empty}
\pagestyle{empty}

\begin{abstract}

Learning fine-grained movements is a challenging topic in robotics, particularly in the context of robotic hands. One specific instance of this challenge is the acquisition of fingerspelling sign language in robots. In this paper, we propose an approach for learning dexterous motor imitation from video examples without additional information.
To achieve this, we first build a URDF model of a robotic hand with a single actuator for each joint. We then leverage pre-trained deep vision models to extract the 3D pose of the hand from RGB videos. Next, using state-of-the-art reinforcement learning algorithms for motion imitation (namely, proximal policy optimization and soft actor-critic), we train a policy to reproduce the movement extracted from the demonstrations. We identify the optimal set of hyperparameters for imitation based on a reference motion.
Finally, we demonstrate the generalizability of our approach by testing it on six different tasks, corresponding to fingerspelled letters. Our results show that our approach is able to successfully imitate these fine-grained movements without additional information, highlighting its potential for real-world applications in robotics. 

\end{abstract}

\section{Introduction}

In recent years, robots have been increasingly utilised across a variety of settings, ranging from industrial to domestic applications. As a result, there is a growing need to enable robots to perform a wide range of actions efficiently. While one approach is to hard-code as many primitives as possible, this strategy requires significant resources and is limited by the difficulty of anticipating all possible variations in actions. Another approach involves teleoperation, which necessitates a human controller and an array of sensors~\cite{icub_teleop} to capture different movements. A more practical and efficient solution involves observing a tutor performing a given action and learning to reproduce it, similar to the way in which people acquire new skills. For instance, one notable example is the \textbf{acquisition of robotic sign language}.

Sign language is a natural language that utilises visual means (like gestures and facial expressions) to convey messages, in contrast to spoken language that relies on sound. The World Federation of the Deaf estimates that there are over 70 million deaf individuals worldwide~\cite{un}, and the World Health Organization predicts that by 2050, nearly 2.5 billion people will have some degree of hearing loss and at least 700 million will require hearing rehabilitation~\cite{who}. 
Recently, researchers tried to address the problem of sign language processing and recognition from a linguistic perspective~\cite{tavella2022phonology}~\cite{tavella2022phonology} rather than a computer vision one, there is not much research on sign language acquisition from a computation point of view, especially in robotics.
A robot that knows such language can significantly enhance human-robot interaction for people who are deaf or hard of hearing, opening up new possibilities for assistive robotics in areas such as healthcare, education, and social interaction, much like current approaches can help people with vision-impairment to automatically recognise objects in pictures taken with a smartphone \cite{polonio2018ghioca}. 
Additionally, a sign language-capable robot can also promote inclusivity and diversity in human-robot interactions, fostering a more welcoming and accessible environment for all. While some research has begun to tackle this problem by focusing on imitation or translation~\cite{zhang}~\cite{hosseini2019teaching}, neither of these approaches allows robots to learn a \textbf{embodied neural representation} of different signs using only RGB data, as they concentrate on retargeting rather than learning or require additional hardware. 

Recently, Learning from Demonstrations (LfD)~\cite{Billard2008} or Learning from Observations (LfO)~\cite{BENTIVEGNA2004163} gained huge popularity in the robotics community. For example, Yang et al.~\cite{Yang_Li_Fermuller_Aloimonos_2015} teach a Baxter robot how to cook by ``watching" YouTube videos, or Peng et al.~\cite{peng_dog} make a robotic dog learn how to move by imitating an actual dog. Similarly, ~\cite{deepmimic} and~\cite{sfv} perform virtual character animation by teaching a virtual humanoid different skills based on motion capture data, like walking or performing backflips.

\begin{figure*}[ht]
    \centering
    \includegraphics[width=\textwidth]{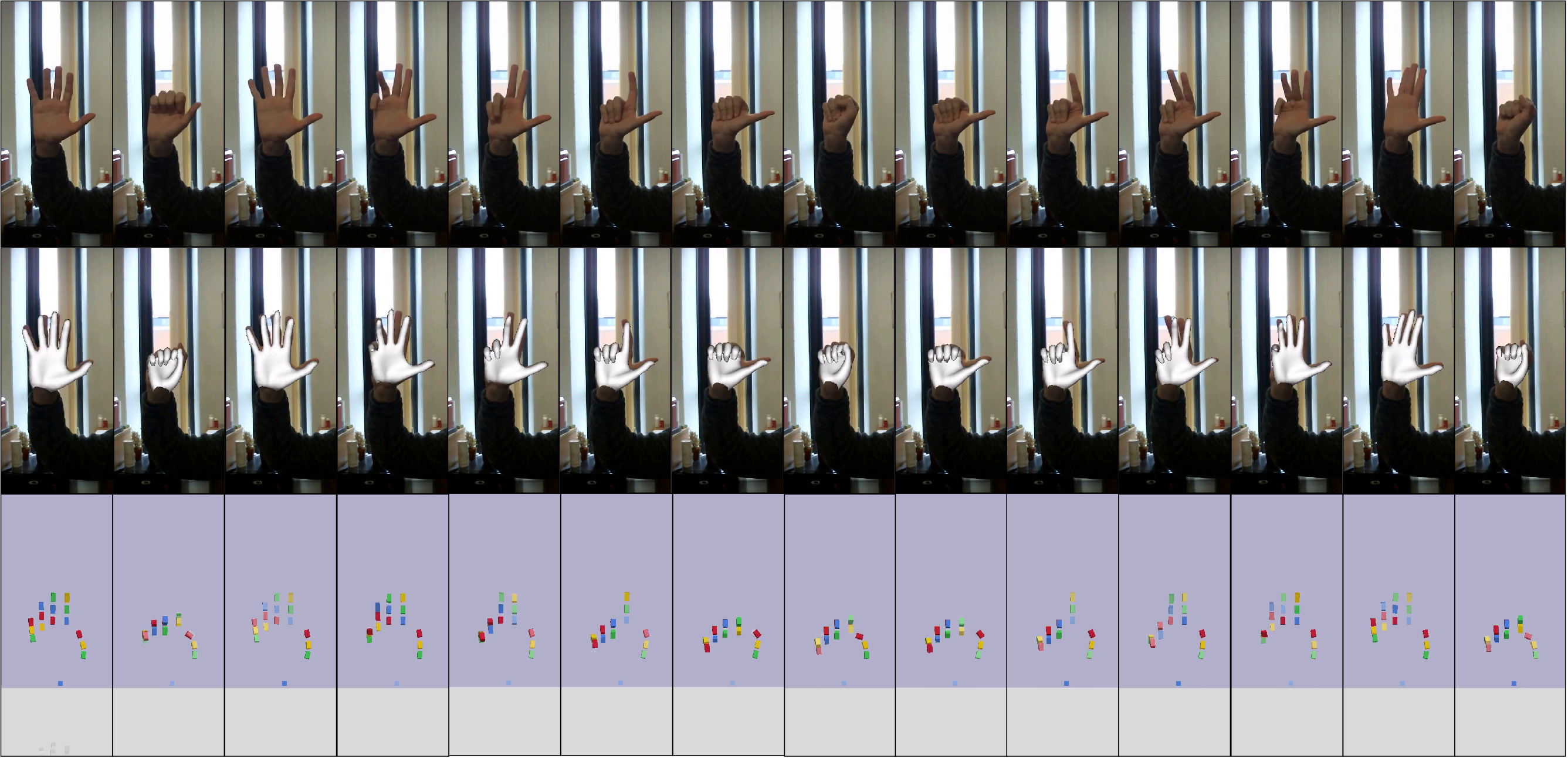}
    \caption{SiLa extracts 3D coordinates and rotations from RGB videos using deep models. It then trains a policy using reinforcement learning, in order to teach to our robotic hand how to imitate the reference motion.}
    \label{fig:intro}
\end{figure*}

We propose an imitation learning approach to fingerspelled American sign language acquisition in robot (namely \textit{SiLa}, short for Signs of Language) based on reinforcement learning, inspired by~\cite{deepmimic}~\cite{sfv}. Our approach is illustrated in Figure~\ref{fig:intro}. Firstly, we build an URDF model of a human hand to simulate an artificial hand using a physics engine~\cite{pybullet} and perform parameter tuning to estimate the controller parameters using a Bayesian approach. Secondly, we extract a 3D mesh of the hand from videos -- which contains 3D coordinates and rotations -- by exploiting a pre-trained vision model. Finally, we use reinforcement learning to estimate a policy which, comparing random movements with the reference motion, enables the simulated hand to imitate the original sign.
To summarise, the contributions of our research are manifold:
\begin{enumerate}
    \item Introducing the problem of sign language acquisition in robots from an embodied perspective,
    \item Providing a new model for a robotic hand, which will be made publicly available on GitHub,
    \item Proving how a Bayesian approach can beused to perform the tuning of a PD controller,
    \item Establishing the reproducibility of DeepMimic~\cite{deepmimic} to a different scenario and with a different learning algorithm, 
    \item Demonstrating the possibility of acquiring fingerspelling skills via reinforcement learning based on RGB video examples,
    \item Creating a new experimental setup to evaluate the performance of the algorithm in the context of skills acquisition from demonstrations, in which researchers can simply capture motion data using any webcam and train the simulated hand to replicate the motion.
\end{enumerate}

\section{Related work}


\subsection{Learning from others}

Behbahani et al.~\cite{vibe} learn models of behaviour from videos of traffic scenes collected from a monocular camera. More interestingly, LfD can be applied to control problems in order to teach virtual characters how to acquire different skills. One of the most relevant works in this direction is the one performed by Peng et al.~\cite{deepmimic}, who build a virtual humanoid character who learns from motion capture data using reinforcement learning. Using this technique, the authors were able to teach a virtual character in a physical environment to perform different actions (e.g., walking, jumping, backflip). In addition, \cite{peng_dog} applies the same approach to a simulated robotic dog instead of a humanoid and includes a domain adaptation part to transfer the policy to a physical robotic dog. \cite{sfv} builds upon~\cite{deepmimic} integrating a video extraction and reconstruction pipeline to learn skills from videos in the wild. Similarly, Lee et al.~\cite{lee2019} teach a simulated humanoid to perform different actions, but using a muscle-actuated human model instead of a motor-actuated one. \cite{vid2game} extract a controllable virtual character from a video of a person performing a certain action, like playing tennis. Finally, \cite{peng_adversarial} overcome the necessity of manually designing imitation techniques for motion selection by using an automated approach which exploits adversarial imitation learning~\cite{gail}.

\subsection{RL and robotic hands}

In addition to humanoid character control, reinforcement learning has been successfully applied to robotic and simulated hands. Most of the recent works use either the MIA hand or the Shadow dexterous hand~\cite{shadowhand}. Saito et al.~\cite{task_grasping} functionally mimic human grasps using a robotic hand and reinforcement learning, develop a grasping reward, and apply domain randomisation to improve policy robustness. However, to perform grasping, it queries a library of grasping skills (i.e., a lookup table), so it actually takes advantage of a set of pre-defined grasping primitives. 
\cite{modular_hand} use a motion capture glove combined with the MANO model~\cite{mano} in order to collect data for objects manipulation. Then they perform optimisation of the finger joints (an inverse kinematics problem) to match the glove and MANO model keypoints. Finally, they transfer movements to the MIA and Shadow dexterous hands and perform simulation on PyBullet~\cite{pybullet}. 
\cite{garcia_hernando} use inverse kinematics to extract rotations from 3D keypoints in order to train a robotic hand. In addition, they exploit two existing datasets -- BigHand2.2M~\cite{bighand} and the one introduced in~\cite{Rajeswaran} (captured with a haptic glove) -- with ground truth in order to introduce an additional reward that encourages the policy to produce actions that resemble the user pose. 
Radosavovic et al.~\cite{radosavovic} propose state-only imitation learning (as opposed to standard state-action imitation learning) to address the lack of demonstration from which to learn skills, allowing learning from demonstrations coming from a different but related setting. \cite{dexmv} propose a novel demonstration-based method to convert human motion to robot demonstrations by extracting from videos 3D object and hand poses. However, while they exploit the MANO model, they use off-the-shelf skin segmentation and hand detection models to obtain a hand mask and an additional 2D pose estimator to extract the hand 2D keypoints and MANO parameters. Finally, they formulate the 3D hand joint estimation as an optimisation problem. They perform motion retargeting, while the fact that we predict rotations rather than keypoints means we do not require retargeting. 

\subsection{Robotic sign language acquisition}
Alabbad et al.~\cite{alabbad} train a Pepper robot to recognise Arabic fingerspelled Arabic sign language letters and use natural language processing to translate written or spoken Arabic to sign language.
Gago et al.~\cite{Gago2019SequencetoSequenceNL} perform natural language to sign language translation in a human-robot interaction scenario. They combine sequence-to-sequence models to a lookup table to make a robot perform a series of signs -- through retargeting and inverse kinematics -- based on a corresponding written input sequence. Alternatively, Zhang et al.~\cite{zhang} use latent optimization to perform retargeting of Chinese sign language on three different virtual robots: YuMi, NAO, and Pepper. Hosseini et al.~\cite{hosseini2019teaching} teach to a RASA robot 10 different Persian signs using a mocap suit, mapping the user gestures to the robot joint space via interpolation and training a set of parallel Hidden Markov Models to encode each sign. This work is the one that resembles the most our approach, but it relies in additional hardware to capture signs. Moreover, it requires the demonstrator to repeat the sign until they are satisfied with the motion mirrored by the robot, introducing a possible bias regarding data acquisition (i.e., repeating the sign until the gesture performed by the robot is understandable).

\section{Methods}

Our study is composed of a series of stages, each of which is detailed in the following paragraphs. Initially, we present the development of our hand model, outlining the technical specifications involved in its creation. Subsequently, we describe the utilisation of a deep learning model to extract pertinent information from RGB videos, which is then exploited for motion emulation. Following that, we elucidate the implementation of reinforcement learning techniques for motion imitation. Finally, we provide a comprehensive overview of the problem at hand, framed in the context of a reinforcement learning problem.

\subsection{Hand model}

\begin{figure}[ht]
        \centering
        \includegraphics[width=0.44\textwidth]{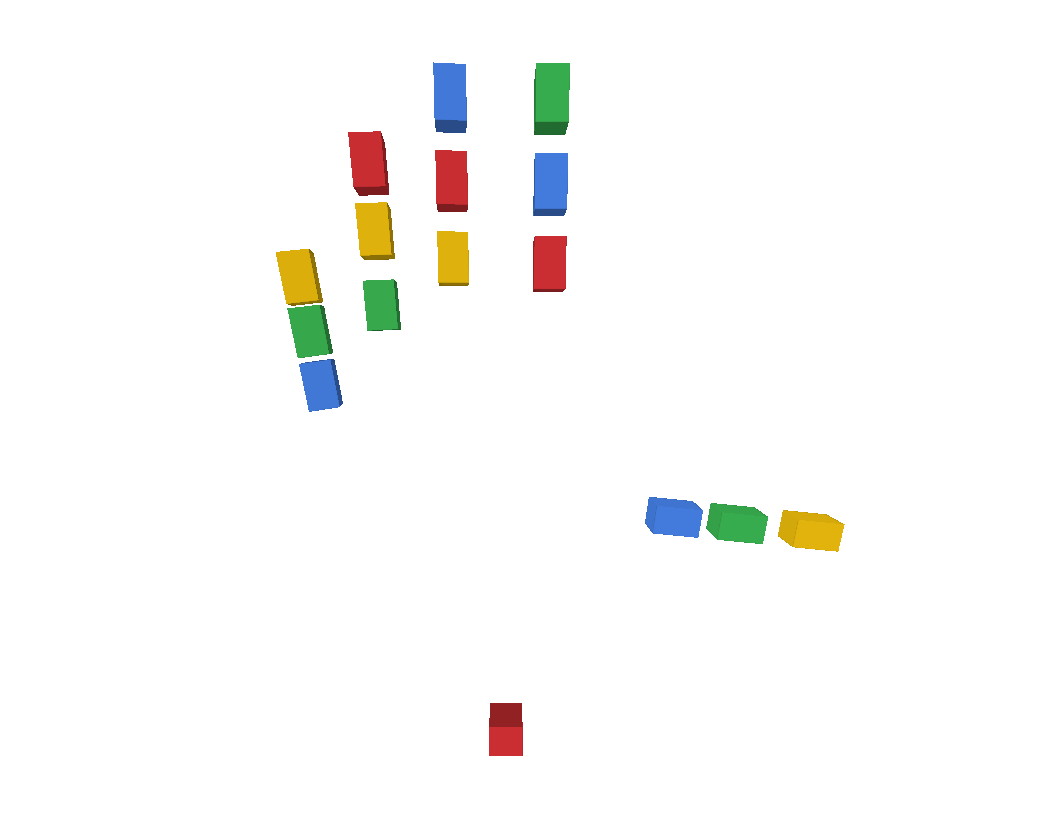}
        \label{fig:hand_urdf}
    \caption{URDF model of our robotic hand.}
    \label{fig:hand_model}
\end{figure}

We build a model of a robotic hand based on the properties of a real human hand. To do so, we measure the position of the different joints of the hand and create an Unified Robotics Description Format (URDF) model with the joints positioned accordingly. Moreover, we impose realistic angular limitations to the joint of each finger. 
Our robotic hand is made of 5 fingers, where each finger is made of 3 joints. For our purposes, we do not consider the wrist as a mobile joint as we focus on movements of the fingers. In addition, we limit the movements of each finger joints to a single axis, as we believe it to be a reasonable approximation -- for this specific task -- about how a human hand works. Thus, in total the model has 15 degrees of freedom (DoF). We impose a range limit for all the joints between $[0, 2]$, where $0$ radiant correspond to a fully-opened hand and $2$ radiant (i.e., slightly more than $\pi/2$) to the joint fully bent. We simulate the robotic hand using PyBullet~\cite{pybullet}, an open source physics simulator. For each joint motor, the simulated PD controller calculates the error as 
$
    \varepsilon = k_p \Delta P + k_d \Delta V
$
where $k_p$ and $k_d$ are, respectively, the position and velocity gains, and $\Delta P = P - \hat{P}$ and $\Delta V = V - \hat{V}$ are the position and velocity errors (i.e., the difference between the desired and the expected value).
The only difference between our robotic hand and a human hand are the degrees of freedom and how the joints move. While in a finger, the joints are not independent due to tendons (i.e., flexing one joint will cause other joints to flex as well due to tendons), the joints in our robotic hand can move independently from each other. This significantly simplifies the modelling of the hand but it adds an ulterior layer of complexity to the imitation task. Figure~\ref{fig:hand_model} illustrates the final URDF model of our hand.

\subsection{Motion extraction}

FrankMocap~\cite{frankmocap} is a state-of-the-art system for 3D motion capture. It uses a deep learning model to predict the 3D motion of a person from a single RGB image. We chose to use FrankMocap for this task because it is easy to use and it is capable of accurately tracking the motion of a person.
It takes an RGB image as input and outputs a 3D mesh based on the SMPL/SMPL-X models~\cite{smpl}~\cite{smplx}. In addition, it also provides 3D keypoints and joint rotations for both body and hands. We use this pre-trained model to extract joint rotations for each finger. As mentioned above, the joints of our robotic hand are capable of rotating along a single axis. Consequently, as the output of FrankMocap is expressed in axis-angle format, we discard the 2 additional angles.

\subsection{Motion imitation}

Reinforcement learning (RL)~\cite{sutton} is commonly formulated as a Markov Decision Process (MDP)~\cite{markov}, which has four main components: a set of states $S$, a set of actions $A$, a reward function $R$, and a policy $\pi$. In brief, a policy can be interpreted as the strategy of the agent to solve the task. A learning agent observes the environment and its own state $s$, and based on these observations (i.e., state) takes an action $a$ to transition to a new state $s'$ based on a probability 

\begin{equation}
    P_{a}(s, s') = Pr(s_{t+1} = s' | s_t = s, a_t = a)
\end{equation}

which yields to a reward $R_{a}(s, s')$ (i.e., an indicator of how good/bad was the chosen action). The final objective is to learn a policy -- a mapping composed of actions chosen by the agent -- which maximises the expected cumulative reward. Additionally, if a policy is parametric ($\pi_{\theta}$), it has to discover the optimal parameters $\theta^*$ that maximise the expected cumulative reward (i.e., the sum of the rewards at each different steps).
There are several algorithms that can be used to find an optimal policy. Such algorithm is usually chosen based on the type of action and/or state spaces (discrete or continuous).
For our scenario, we use the Proximal Policy Optimisation (PPO)~\cite{ppo} and Soft Actor-Critic (SAC)~\cite{sac} algorithms to estimate a policy for our problem.

PPO~\cite{ppo} is an optimisation algorithm used when both action and state spaces are continuous. It is a policy gradient method, where the gradient of the expected cumulative reward is calculated using trajectories $\tau$ -- i.e., sequences of $(s, a, r)$ over a set of contiguous time steps -- sampled by following the policy. Thus, given a parametric policy $\pi_{\theta}$ and $T$ steps, the expected reward is

\begin{equation}
    J(\theta) = \mathbb{E}_{\tau \sim p_{\theta}(\tau)} \left[ \sum_{t=0}^{T} \gamma^t r_t \right]
\end{equation}

where $p_{\theta}(\tau) = p(s_0) \prod_{t=0}^{T-1} p(s_{t+1} | s_t a_t) \pi_{\theta}(a_t | s_t)$ is the distribution over all possible trajectories $\tau = (s_0, a_0, s_1, a_1, ..., a_{T-1}, s_{T})$ induced by the policy $p_\theta$, $p(s_0)$ being the initial state distribution and $\gamma \in [0, 1]$ is a discount factor used to ensure that the reward has an upper bound. The policy gradient can be estimated as

\begin{equation}
    \nabla_\theta J(\theta) = \mathbb{E}_{s_t \sim d_{\theta}(s_t), a_t \sim \pi_{\theta}(a_t | s_t)} \left[ \nabla_\theta log(\pi_\theta(a_t | s_t)) \mathcal{A}_t \right]
\end{equation}

where is $d_{\theta} (s_t)$ is the distribution of states under the policy $\pi_\theta$, while $\mathcal{A}_t = R_t - V(s_t)$ represents the advantage of taking an action $a_t$ from a given state $s_t$. $R_t$ is the reward by a particular trajectory starting from state $s_t$ at time t, and 

\begin{equation}
    V(s_t) = \mathbb{E} \left[ R_t | \pi_\theta, s_t \right]
\end{equation}

is the value function that estimates the average reward for starting at $s_t$ and following the policy for all subsequent steps.

SAC~\cite{sac} is a reinforcement learning method designed to learn policies for continuous control tasks. SAC is based on the principle of maximum entropy reinforcement learning, where the training of the policy involves finding the right balance between expected return and entropy, which is an indicator of the level of randomness of the policy. This relationship has a strong link to the exploration-exploitation trade-off, whereby boosting entropy can lead to increased exploration and subsequently, faster learning. 
The core idea of SAC is to use a soft value function, which estimates the expected cumulative reward while incorporating an entropy bonus term $H$. The entropy bonus encourages exploration and ensures that the policy remains diverse and adaptive. The soft Q-function is defined as:

\begin{equation}
    Q^\pi(s, a) = E_{s' \sim P} [R(s, a, s') + \gamma V^\pi(s')]
\end{equation}

where $R(s, a, s')$ is the reward received after taking action $a$ in state $s$ and ending up in state $s'$, $\gamma$ is the discount factor, $V$ is the soft state-value function, and $\pi$ is the policy.
The soft state-value function is learned using an off-policy actor-critic algorithm, which uses a separate policy network to select actions. The soft state-value function is defined as:

\begin{equation}
    V(s) = E_{a \sim \pi} [Q(s, a)] + \alpha  H(\pi(\cdot|s))
\end{equation}

where $Q$ is the soft Q-function, $\pi$ is the policy, $\alpha$ is the temperature parameter controlling the trade-off between exploration and exploitation, and $H$ is the entropy of the policy.
For more details about PPO and SAC, we redirect the reader to their respective papers \cite{ppo} and \cite{sac}.

\subsection{Problem statement}

We formulate the control problem of a robotic hand as a MDP. The action space is composed of different poses (i.e., combination of joint positions) for the controller, while the state describes a configuration of the hand, position and linear/angular velocity for each component of the hand. Moreover, a phase component $\phi \in [0, 1]$ is added to the state space to synchronise the target and reference motions. We represent the policy using a multilayer perceptron with 2 hidden layers, which input and output dimensions are dictated by the space and action size. The number of hidden units is 1024 and 512 respectively, based on~\cite{deepmimic}. We estimate the pose error by calculating the scalar rotation of the quaternion difference between the simulated hand and the reference motion. In practice, this is similar to the mean squared error applied to quaternions but in the quaternion space rather than the euclidean one. At time $t$, given the desired velocity $v_{j,t}$ and the simulated velocity $\hat{v}_{j,t}$ for joint $j$, the velocity error is

\begin{equation}
    \varepsilon_t^v = \sum_{j} ||\hat{v}_{j,t} - v_{j,t}||^2.
    \label{eq:error}
\end{equation}

Additionally, we calculate an end effectors (i.e., fingertips) error $\varepsilon_t^{e}$ and a root error $\varepsilon_t^{r}$ in the same manner. The former ensures that the 3D world position (in meters) of fingertips correspond, while the latter penalises deviations from root orientation when compared to the reference motion. Finally, the reward is calculated on the basis of the errors as follows

\begin{equation}
    r_t = w^p r_t^p +  w^v r_t^v +  w^e r_t^e + w^r r_t^r
\end{equation}

where $w^x$ is a weight manually chosen (with the only condition of the different $w^x$ summing to 1) and $r_t^x$ is the reward at time-step $t$ for the component $x$, calculated as $r_t^x = e^{-k^x \varepsilon_t^p}$ with $k^x$ being a factor used to balance the reward based on the magnitude of the error. All the values for $w$ and $k$ are taken from~\cite{deepmimic}. As a proof-of-concept for our proposal, we train a policy for each different task. While it is not as efficient as training a single policy over different tasks, it enables us to easily evaluate the feasibility of our approach.

\section{Experimental setup}

We carry out three major experiments: one to tune the controller, one for searching hyperparameters, and another for training on the actual data. We use Weights and Biases~\cite{wandb} to carry out both the experiments for controller tuning and hyperparameters search. In particular, we use a Bayesian approach instead of a grid search. In this way, the hyperparameters for the $N$ experiment are chosen on the basis of the results of the previous $N-1$ runs. Additionally, we use Stable Baselines 3~\cite{stable-baselines3} implementation of PPO and SAC.

\subsection{Controller tuning}

Our controller can be tuned using two variables, namely $k_p$ and $k_d$. Firstly, we generate a random reference motion which we use as a baseline for the tuning. Secondly, we create a function to minimise, which purpose is to make the reference motion and the simulated hand motion as similar as possible. We define this function as the sum of the pose and velocity errors $\varepsilon_t^p$ and $\varepsilon_t^v$
$
    \varepsilon^{control} = \varepsilon_t^p + \varepsilon_t^v
$
where $\varepsilon_t^p$ and $\varepsilon_t^v$ are defined according to Equation~\ref{eq:error}.
This is because we want the simulated motion to resemble as much as possible the reference one, both in terms of \textit{how} (i.e., pose) and \textit{when} (i.e., velocity) the action is executed. Finally, we use a Bayesian optimisation strategy to find values that minimise $\varepsilon^{control}$.
To reduce the number of different simulations, we run 3 different swipes, characterised by different maximum values for the parameters $k_p$ and $k_d$. The three different upper bounds are 100, 10 and 1 for both parameters.

\subsection{Motion imitation}

Model selection and hyperparameters tuning are two fundamental steps in deep and reinforcement learning.
However, due to the amount of resources necessary to perform a single training instance (approximately between 8 and 14 hours with 1 NVIDIA RTX 2080Ti GPU and 8 cores, depending on the parameters and the algorithm), we opt for a hierarchical approach. 

In the first instance, we explore a subset of hyperparameters in order to understand how they affect training speed and performance. We acquire a reference motion used for the sole purpose of tuning, which is illustrated in Figure~\ref{fig:intro}. Table~\ref{tab:hyperparams} lists all the different values we test during our hyperparameters tuning for PPO. Oppositely, for SAC we use the default hyperparameters from \cite{stable-baselines3} as preliminary experiments have shown they can achieve good results. At evaluation time, we run each policy for 2000 steps, meaning the maximum reward we can achieve is 2000 as the single step reward can be at maximum 1.
Finally, we test the ability of our tuned model to generalise by training different networks - with the set of best hyperparameters identified during the previous step - over 6 different reference motions (i.e., fingerspelled letters A, B, C, D, E, F). We repeat each training session 10 times using different random seeds, in order to ensure statistical significance.

\begin{table}[ht]
\centering
\begin{tabular}{ll}
\hline \hline
\textbf{Parameter}    & \textbf{Values}                        \\ \hline
learning rate         & (1, 3, 10, 30) x $10^{-6}$  \\
number of steps               & 512, 1024, 4096                 \\
weight decay rate     & (1, 10) x $10^{-5}$                 \\
batch size            & 128, 256, 512                 \\
discount factor       & 0.9, 0.95           \\
log std dev           & -3, -2, -1                             \\

number of epochs              & 3, 5, 10                 \\ \hline \hline
\end{tabular}
\caption{Values of different hyperparameters explored during tuning for PPO.}
\label{tab:hyperparams}
\end{table}

\section{Results and discussion}

\subsection{Controller tuning}

We describe the results (providing Pearson correlation coefficient PCC~\cite{freedman2007statistics} between the error and the parameters) and graphically report the most relevant sweep.

\begin{figure}[ht]
    \centering
    \includegraphics[width=0.49\textwidth]{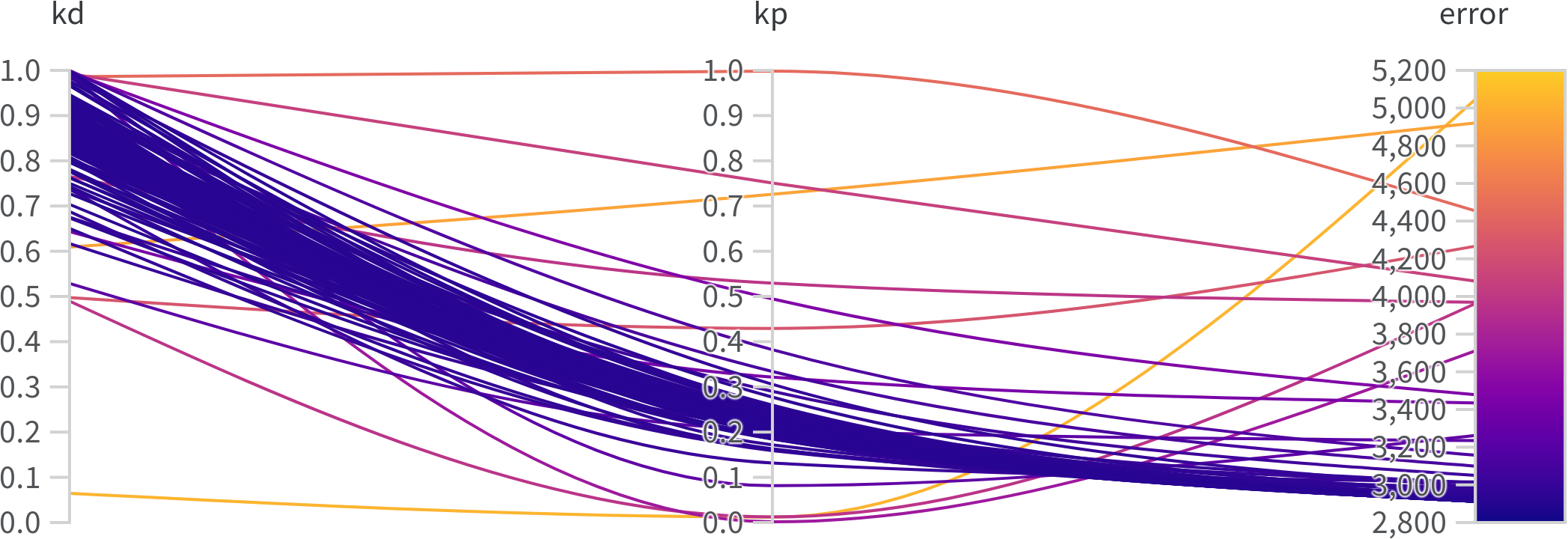}
    \caption{Error for $k_p$ and $k_d$ with maximum values equal to 1}
    \label{fig:controller1_squish}
\end{figure}

Our first sweep is the one with the largest range: [0-100]. On one hand, there is no clear combination of values, which leads to a small error. However, it is noticeable that the runs leading to the lowest error are those in which $k_d < k_p$ (PCC 0.45 for $k_d$ and -0.44 for $k_p$), but the error is very high. On the other hand, the sweep with the range [0, 10] clearly indicates that the values leading to the minimum error are the ones in the range [0, 2] (PCC 0.71 for both $k_d$ and $k_p$). Additionally, the error is almost one order of magnitude smaller than the previous sweep (86K vs 10K). Finally, Figure~\ref{fig:controller1_squish} gives us a compromise between the observations we made using the previous sweeps: the best results are obtained when $k_d > k_p$ and with small values (PCC -0.49 for $k_d$ and 0.51 for $k_p$). 

\begin{figure}[ht]
    \centering
    \includegraphics[width=0.48\textwidth]{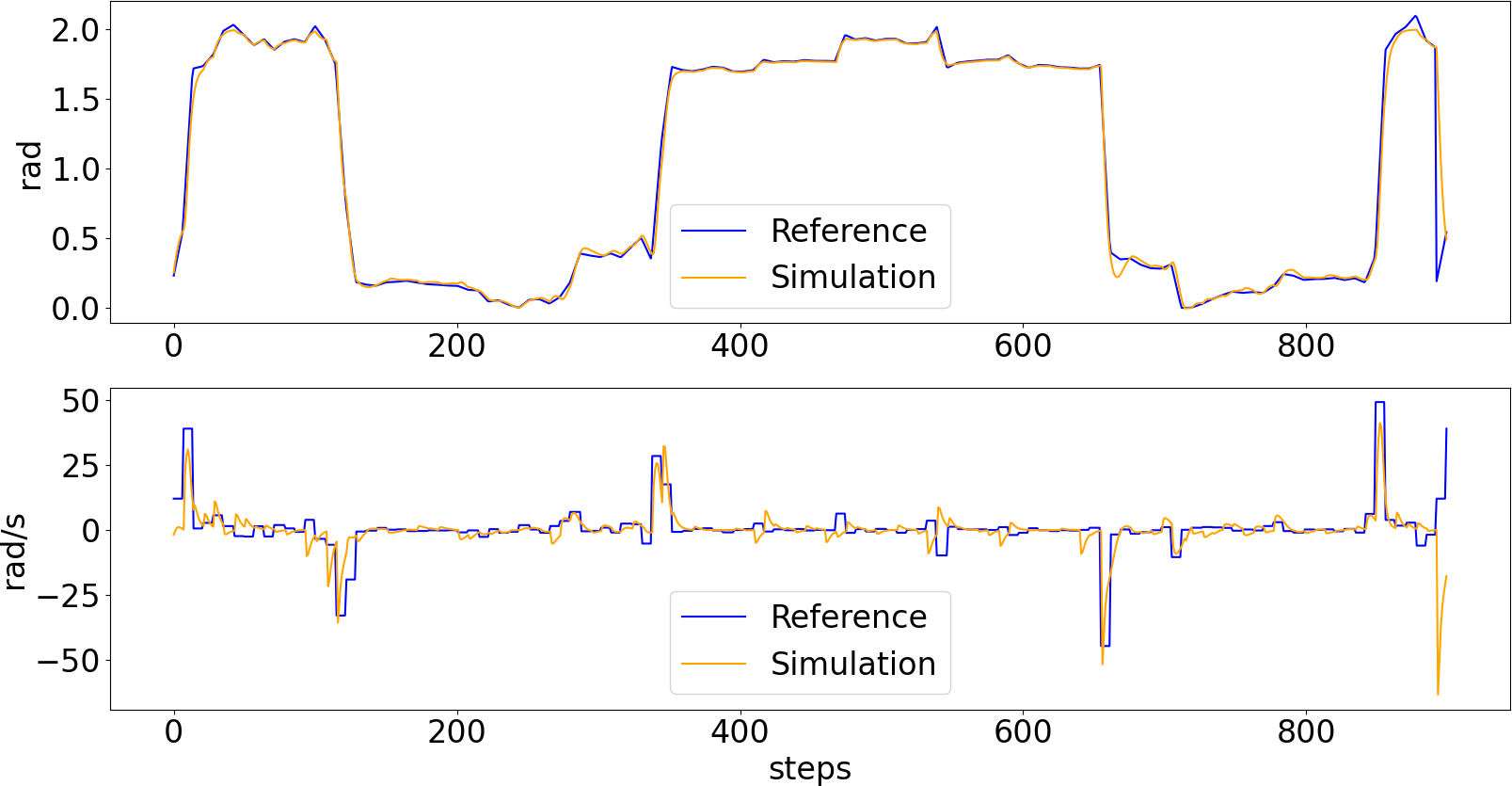}
    \caption{Comparison between the reference and simulated position (top) and velocity (bottom) for the last  phalanx of the index finger using the best couple of parameters.}
    \label{fig:tuned}
\end{figure}

Finally, Figure~\ref{fig:tuned} shows a comparison of the position and velocity of the reference and simulated motions of the last phalanx of the index finger, using the best values $k_d = 0.87$ and $k_p = 0.22$.

\subsection{Motion imitation}

\begin{figure}[ht]
    \begin{subfigure}{.47\textwidth}
        \centering
        \includegraphics[width=\textwidth]{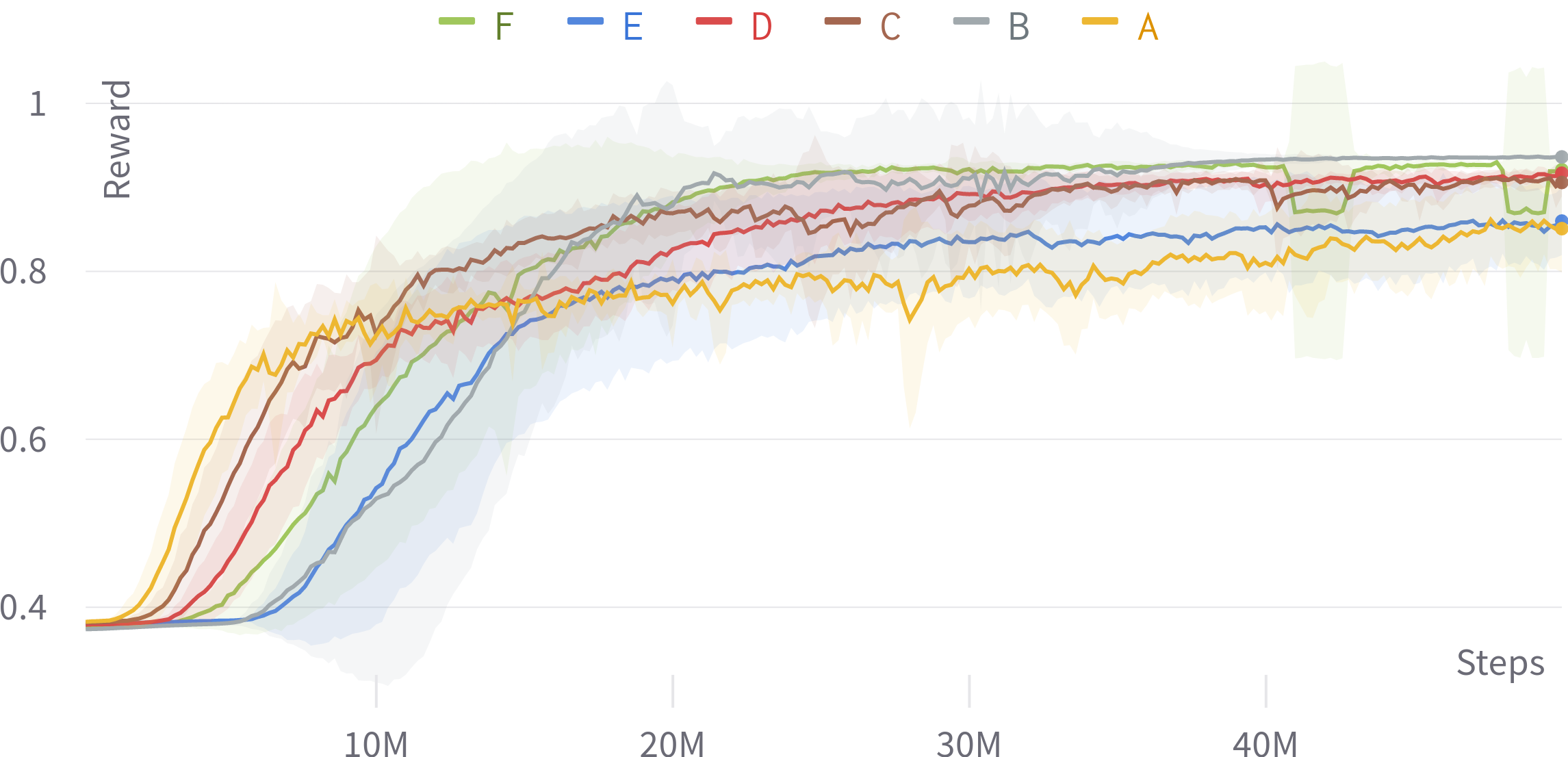}
        \caption{PPO}
        \label{fig:imitation_ppo}
    \end{subfigure}
    \begin{subfigure}{.47\textwidth}
        \centering
        \includegraphics[width=\textwidth]{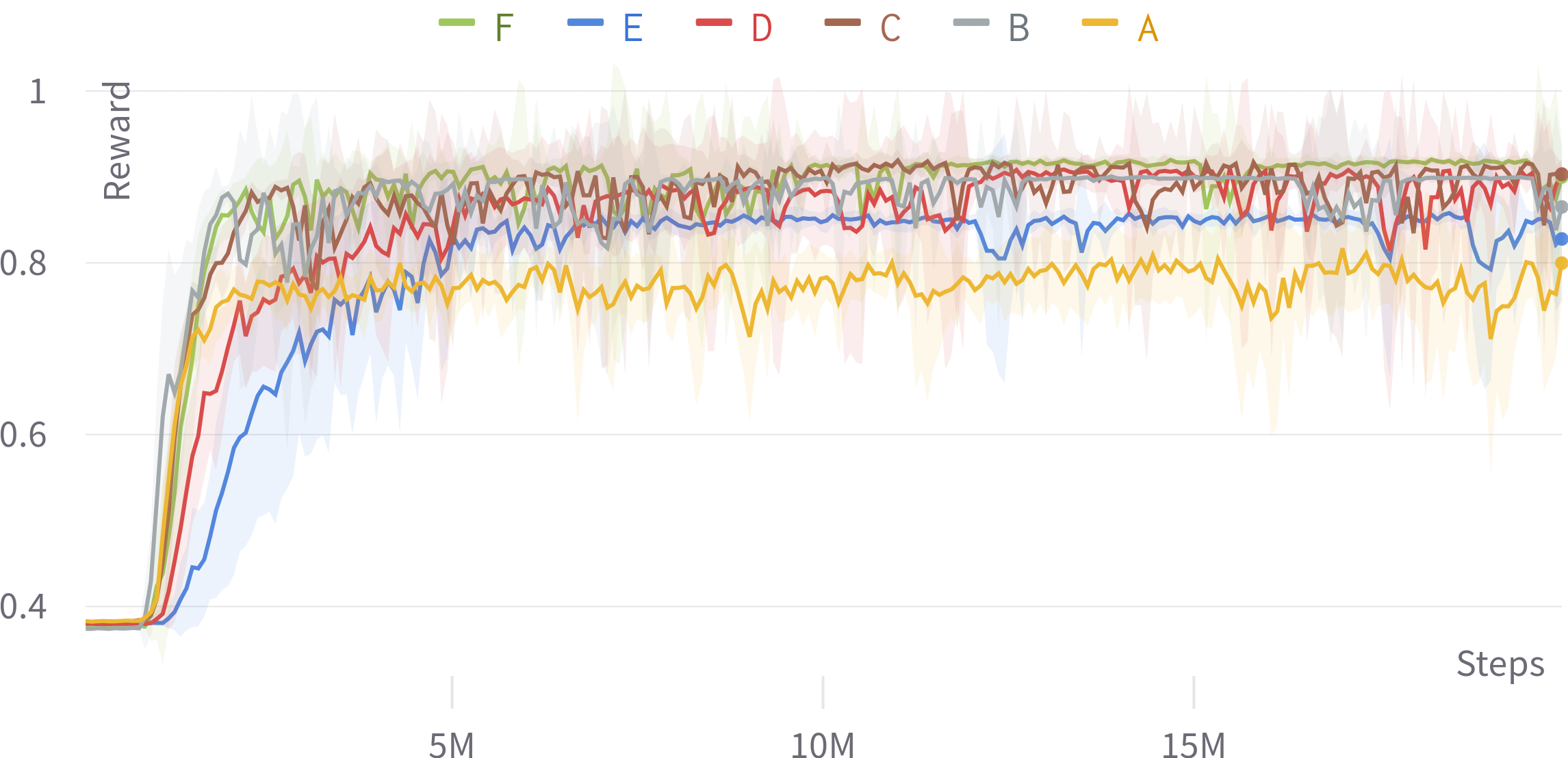}
        \caption{SAC}
        \label{fig:imitation_sac}
    \end{subfigure}
    \caption{Single step reward increase over different phases of training. Notice that the maximum number of steps for PPO is 50 millions, while for SAC is 20 millions.}
    \label{fig:imitation}
\end{figure}

During our experiments for PPO, we found that no particular single hyperparameter contributed significantly to a high reward. However, it is clear that there is a huge difference between good and bad combinations of hyperparameters. In fact, we found out that the worst run (which leads to a mean reward of 854) uses the following values: $batch\_size=128$, $gamma=0.9$, $learning\_rate=10^{-5}$, $log\_std\_init=-3$, $n\_epochs=10$, $weight_decay=10^{-5}$ and $n\_steps=1024$. On the opposite, the best run (mean reward equal to 1624) uses $batch\_size=128$, $gamma=0.9$, $learning\_rate=10^{-5}$, $log\_std\_init=-2$ and $n\_epochs=10$, $weight_decay=10^{-5}$ and $n\_steps=1024$. Hence, the only difference between the two runs is the value of the $log\_std\_init$ parameter. Nevertheless, when we calculate the PCC between this parameter and the reward, we obtain a value of 0.081, indicating no clear linear correlation between the two values.
Last but not least, we train our policies to imitate six different fingerspelled letters using the previously identified hyperparameters. Figure~\ref{fig:imitation} illustrates the results of our training over 10 different seeds for PPO (Figure~\ref{fig:imitation_ppo}) and SAC (Figure~\ref{fig:imitation_sac}). All policies have a minimum reward of 0.4 due to the normalised centre-of-mass reward being always 1, given that the hand as a whole is not moving from its initial position. Additionally, all the policies converge at most after 50 million steps. The single step values for the reward $r_t$ vary between 0.8 and 0.95 at the end of the training process. Hence, we conclude that our model is able to learn different motions using the same model and hyperparameters. However, based on the learning curves, we can see how PPO produces less fluctuations in the reward, while SAC struggles to fully converges and the reward continues fluctuating. Despite this shortcoming, SAC learns much faster than PPO does (20 million vs 50 million steps), providing a more efficient alternative to PPO which yields to comparable results. Finally, Table~\ref{tab:final} provides a summary of the performance of each learnt policy when compared to motion retargeting. We can see how, in most cases, the performance of our learning approach achieves comparable performance to simple retargeting, with the advantage of an actual learning process that creates an embodied neural representation of the different signs.

\begin{table}[ht]
\centering
\begin{tabular}{cccc}
\hline \hline
\textbf{Letter} & \textbf{Retargeting} & \textbf{PPO}   & \textbf{SAC}   \\ \hline
A               & \textbf{1905}                 & \underline{$1700 \pm 106$} & $1661 \pm 62$  \\
B               & \textbf{1941}                 & \underline{$1920 \pm 35$}  & $1878 \pm 186$ \\
C               & \textbf{1899}                 & $1833 \pm 37$  & \underline{$1873 \pm 34$}   \\
D               & 1876                 & $1828 \pm 32$  & \underline{\bm{$1887 \pm 19$}}  \\
E               & \textbf{1915}                 & $1705 \pm 87$  & \underline{$1803 \pm 98$}  \\
F               & 1915                 & $1893 \pm 38$  & \underline{\bm{$1929 \pm 57$}}  \\ \hline \hline
\end{tabular}%
\caption{Cumulative rewards of PPO and SAC for each different letter over 10 different seeds. Highest score in bold, best learning algorithm underlined.}
\label{tab:final}
\end{table}

\section{Conclusion and future work}

Our research focuses on the embodied acquisition in robots of sign language fingerspelling through imitation learning from RGB videos. This is a challenging task, as it requires the imitation of fine-grained movements. We developed an URDF model of a robotic hand, identified the parameters for the hand PD controller using a Bayesian approach and the hyperparameters for the imitation algorithm. Moreover, we compare two different learning algorithms (PPO and SAC) over the same task. In the end, we achieve imitation over 6 different fingerspelled letters with performance comparable to ideal retargeting.
As a first approach, we chose to limit the DoFs to 1 for each joint. However, while this is correct for the distal and the proximal interphalangeal joints (i.e., the two outermost joints), this is not the case for the metacarpophalangeal joint. In particular, this constraint limits the mobility of the thumb. In addition, we limited wrist mobility, which is important for dexterous movements such as the one involved in sign language. All these physical limitations will be addressed in the next stage of our research by building upon our current model.
Additional limitations are related to the learning algorithm we chose, and how we use it. First and foremost, reinforcement learning is known for being computationally expensive. In fact, we previously mentioned how a single run requires several hours. This shortcoming becomes even more relevant when we consider that we train a different policy for each different motion, as opposed to more efficient approaches that take advantage of motion priors~\cite{peng_adversarial} or mixture-of-experts. Nevertheless, recent progress in GPU-based simulations (e.g., NVIDIA Isaac sim~\cite{isaac}) and motion priors~\cite{peng_adversarial} can significantly speed up the learning process. 
As future steps, we envision a new robotic hand with additional degrees of freedom, capable of imitating more complex motions. In addition, we plan to expand the evaluation to cover the entire fingerspelled alphabet and explore more efficient methodologies, such as mixture-of-experts~\cite{won} or motion priors~\cite{peng_adversarial}. SiLa could not only be used for scenarios in robotics (e.g., generation of grasping poses) but also for generating realistic animations for virtual characters. In the future, we envision SiLa as tool for deaf people to communicate with robots in a way that is natural and intuitive for them, allowing for more seamless and inclusive interactions.



\bibliographystyle{ieeetr}
\bibliography{root.bib}


\end{document}